\documentclass[9pt,a4paper,twocolumn,twoside]{rho-class/rho}
\usepackage[english]{babel}
\newcommand{\sparkcase}{G1SportMode\_\allowbreak{}D1\_\allowbreak{}WG\_\allowbreak{}SO\_\allowbreak{}v1}

\doctype{Preprint}
\title{Adversarial Stress Testing of SPARK Humanoid Safety Filters}

\author[{\textasteriskcentered},a,1]{Saurav Ghosh}
\author[a,2]{Abdou Sow}
\author[a,3]{Luke Zhang}

\affil[a]{Department of Computer Science and Engineering, Washington University in St. Louis, Missouri, United States}
\corres{\textsuperscript{\textasteriskcentered}Corresponding author: \href{mailto:saurav.ghosh@wustl.edu}{saurav.ghosh@wustl.edu}}

\docinfo{This manuscript is a preprint and has not been peer reviewed.}
\journalname{arXiv Preprint}
\journal{Preprint}
\theday{\today}
\vol{}
\no{}

\begin{abstract}
Humanoid robots are difficult to deploy safely because they have high-dimensional bodies, many collision constraints, and must operate near people and obstacles. Safety filters help by modifying a nominal control action when it may violate collision-avoidance constraints. Still, nominal benchmark scores do not fully show how these filters behave in harder environments. In this work, we study the robustness of SPARK humanoid safety filters through replication and stress testing. We replicate the SPARK benchmark case \sparkcase{} in MuJoCo and evaluate RSSA, RSSS, SSA, CBF, PFM, and SMA under controlled random seeds. We also built a post-processing pipeline that converts raw SPARK logs into goal-tracking, minimum-distance, and collision-step metrics. Our results show that some methods track the goal more closely, while others reduce collision steps more effectively. The stress tests further indicate that safety behavior can change under obstacle crowding, noisy distance estimates, and delayed obstacle information. These findings suggest that humanoid autonomy should be evaluated beyond nominal performance, using metrics that expose failure modes before deployment.
\end{abstract}

%----------------------------------------------------------

\begin{document}

    \maketitle
    \thispagestyle{firststyle}

%----------------------------------------------------------

\section{Introduction}
\label{sec:introduction}

Humanoid robots operate with many moving links, high-dimensional dynamics, and frequent interaction with nearby people, objects, and obstacles. This makes safety a central requirement for trustworthy autonomy. Safety filters address this problem by modifying a nominal control action when it may violate collision-avoidance constraints.

We study safety-filter robustness using SPARK, a modular benchmark for safe humanoid control~\cite{sun2025}. SPARK supports standardized comparisons across robot configurations, tasks, obstacles, policies, and safety modules. However, nominal benchmark performance alone does not fully capture robustness. A method that works well in a clean benchmark may degrade when obstacle estimates are noisy, obstacle information is delayed, or the environment becomes crowded.

We replicate and analyze the SPARK case \sparkcase{}. We evaluate RSSA, RSSS, SSA, CBF, PFM, and SMA under a shared benchmark setting, fixed time horizon, and controlled seeds. We also built a post-processing pipeline that converts high-dimensional .npz logs into goal-tracking, minimum-distance, and collision-step metrics. Our baseline results show a safety--performance trade-off in this benchmark case: PFM tracks the goal closely but has more collision steps, SMA gives the lowest average environment-collision count, and SSA/RSSA/RSSS show more balanced behavior. We also observe long runtimes and repeated ``No Solution'' outputs, which suggest feasibility limits when constraints become difficult.

This paper makes three main contributions:
\begin{enumerate}
    \item We replicate a SPARK humanoid safety benchmark in MuJoCo and compare six safety filters under a shared benchmark case, fixed simulation horizon, and controlled random seeds.

    \item We build a parsing and post-processing pipeline for SPARK .npz outputs, converting raw simulation logs into final goal distance, minimum environment distance, and collision-step counts.

    \item We stress-test the filters using obstacle crowding, perception noise, and sensor latency, showing that nominal safety--performance behavior can change under harder sensing and environment conditions in this benchmark setting.
\end{enumerate}

The rest of the paper is organized as follows. Section~\ref{sec:preliminaries} reviews safety-filter background and related robustness-testing work. Section~\ref{sec:method} describes the replication pipeline, data extraction process, post-processing metrics, and adversarial stress-test design. Section~\ref{sec:experiments} presents the baseline replication, obstacle-crowding, perception-noise, and latency results. Section~\ref{sec:discussion} discusses the main findings and limitations. Section~\ref{sec:conclusion} concludes with future directions.

\section{Background and Related Work}
\label{sec:preliminaries}

This section summarizes the safety-filter and benchmark concepts used in our experiments.

\subsection{Safety Filters for Robot Control} 

Safety filters are commonly used when a robot has a nominal controller that performs the task. In this setting, the safety module modifies the nominal action only when needed. This idea appears in several forms. Artificial potential fields use repulsive terms to push the robot away from obstacles~\cite{khatib1986real}. Sliding-mode collision avoidance uses the gradient of a safety function to react when a constraint boundary is approached~\cite{gracia2013reactive}. Safe Set Algorithm (SSA) methods define a safety index and choose actions that keep the state inside a safe set~\cite{liu2014safe}. Control Barrier Functions (CBFs) provide a related optimization-based framework for enforcing forward invariance of safe sets in control-affine systems~\cite{ames2019control}. Energy-function-based safe control gives a common view of these methods. Wei and Liu organize several filters, including PFM, SSA, CBF, SMA, and SSS, under a unified framework and study their safety--efficiency behavior~\cite{wei2019safe}. This framework is relevant here because SPARK builds on the same family of safety-filter ideas.

\subsection{Humanoid Safety and Whole-Body Constraints} 

Humanoid robots make safety filtering harder than many lower-dimensional robots. Prior work has used CBFs for bipedal walking and navigation, including safety-critical footstep control and discrete-time bipedal navigation~\cite{nguyen2015safety,agrawal2017discrete}. More recent humanoid work applies CBFs to whole-body controllers for self-collision avoidance and task-space safety~\cite{khazoom2022humanoid,paredes2024safe}. These methods show that formal safety constraints can be integrated into complex humanoid controllers. At the same time, humanoid safety remains difficult because many constraints can be active at once. Reachability-based methods provide another formal view of safety verification for nonlinear and hybrid systems, although scalability remains difficult for high-dimensional systems~\cite{Chen2018HamiltonJacobiRS}. Learning-based safe-control methods and learned certificates also address parts of this problem, but they still require careful validation before deployment~\cite{brunke2022safe,dawson2023safe}. This motivates benchmark-based evaluation, in which different safety methods can be tested across shared tasks and metrics.

\subsection{Benchmarks and Robustness Testing} 

Several benchmarks support research on safe learning and safe control. Safe-Control-Gym provides reinforcement learning in robotics~\cite{yuan2022safecontrolgym}. Safety-Gymnasium provides a broader safe reinforcement learning benchmark suite~\cite{ji2023safetygymnasium}. These benchmarks are useful for comparing algorithms, but they are not specifically focused on humanoid safety filters. SPARK~\cite{sun2025} directly targets this gap by providing a modular benchmark for humanoid safety filters. Our work builds on SPARK by focusing on replication, post-processing of raw outputs, and robustness-oriented stress testing. Nominal benchmarks, however, do not fully explain robustness. This motivates robustness-oriented evaluation, where the goal is not only to report average benchmark performance but also to expose cases where safety behavior degrades. Prior work includes Breach, which supports simulation-based verification and parameter synthesis for hybrid systems~\cite{donze2010breach}. S-TaLiRo provides another related falsification tool for searching low-robustness trajectories in hybrid and Simulink/Stateflow systems~\cite{annpureddy2011s}. VerifAI supports formal design and analysis of AI-based systems, including simulation-guided falsification and scenario exploration under environmental uncertainty~\cite{dreossi2019verifai}. Scenic provides a related scenario-specification language for describing distributions over scenes, which is useful for generating structured test environments for autonomous systems~\cite{10.1145/3314221.3314633}. Simulation-based adversarial testing has also been used for autonomous driving systems with machine-learning components~\cite{tuncali2018simulation}. Adaptive stress testing also frames failure discovery as a search problem, where the goal is to find likely failure scenarios in simulation rather than only measure average-case performance~\cite{koren2018adaptive}.

\section{Methods}
\label{sec:method}

We describe the experimental pipeline used to replicate and stress-test SPARK safety filters. The pipeline has three parts: running a benchmark, parsing simulation logs into metrics, and injecting attacks to evaluate robustness. Our study is intentionally focused on one SPARK benchmark case, \sparkcase{}, rather than the full SPARK benchmark suite. This scope allows us to keep the task, horizon, seeds, and evaluation logic fixed while studying how the same safety filters behave under nominal and stressed conditions.

\subsection{Replication Pipeline} 

We ran the SPARK benchmark in a local Ubuntu-based environment with MuJoCo, the SPARK codebase, and the required Python dependencies. We verified the setup by launching the G1 humanoid benchmark from the command line. All simulation, parsing, and plotting were performed locally using fixed scripts and configurations. The main benchmark entry point was run\_g1\_benchmark.py. We used this script with modified configurations to run the same benchmark across multiple safety filters. The fixed case was \sparkcase{}, which uses the Unitree G1 humanoid in SportMode, first-order dynamics, a whole-body goal, and static obstacles. We selected this case as the common setting so that all methods were evaluated under the same task structure.

\begin{figure}[h]
    \centering
    \includegraphics[width=\linewidth]{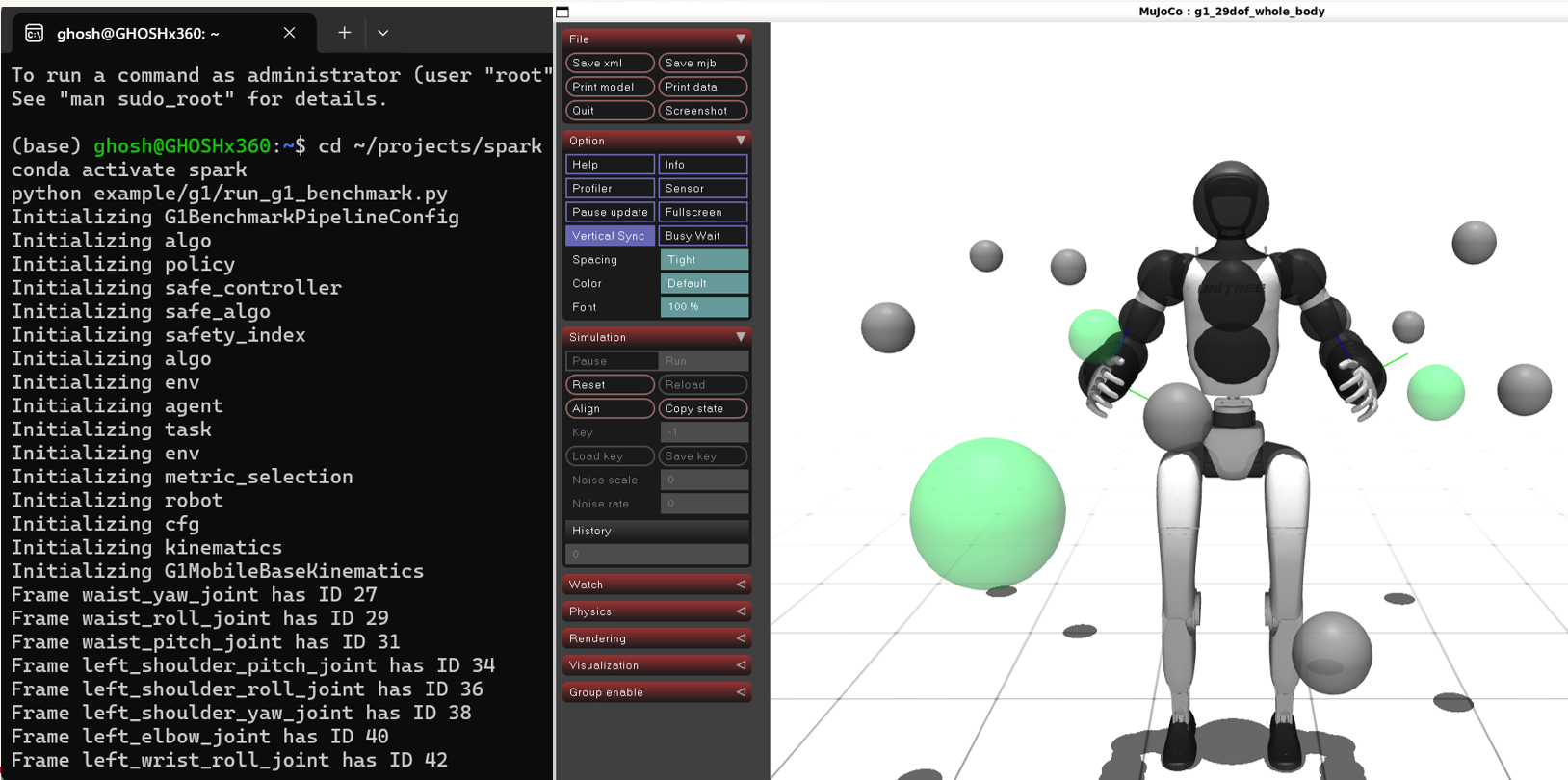}
    \caption{MuJoCo execution view of the replicated SPARK G1 humanoid benchmark. The scene shows the Unitree G1 humanoid, obstacle spheres, goal markers, and terminal-side benchmark initialization.}
    \label{fig:mujoco-g1-setup}
\end{figure}

We evaluated six safety-filter variants: RSSA, RSSS, SSA, CBF, PFM, and SMA. For the baseline replication, we used seeds 20, 21, and 22 with 5000 simulation steps per run, while keeping the task, horizon, and evaluation logic fixed whenever possible. Each completed run produced four main artifacts: \texttt{data.npz}, \texttt{parsed\_metrics.csv}, \texttt{summary.json}, and \texttt{quick\_plot.png}. These outputs support comparison of goal tracking, obstacle distance, collision behavior, and safety--performance trade-offs across methods.

\subsection{Data Extraction and Post-Processing} 

SPARK stores each benchmark run as a compressed \texttt{.npz} file containing step-wise trajectories and metric arrays. Because these files are not directly readable as flat tables, we implemented a post-processing pipeline that parses each \texttt{data.npz} file and exports cleaned metrics to \texttt{parsed\_metrics.csv}. The parser extracts goal-tracking signals such as \texttt{dist\_goal\_arm}, which measures the arm-to-goal distance over time, and safety signals such as robot--environment and self-collision distances. The main safety signal is \texttt{dist\_robot\_to\_env}, a high-dimensional array of distances from the robot's collision volumes to obstacles at each timestep. We reduce this array to one interpretable environment-safety trace:
\begin{equation}
d_{\min}^{\mathrm{env}}(t)=\min_{i,j} d_{i,j}^{\mathrm{env}}(t),
\label{eq:min-env-distance}
\end{equation}
where \(d_{i,j}^{\mathrm{env}}(t)\) is the distance between robot collision volume \(i\) and obstacle \(j\) at timestep \(t\). This trace captures the closest robot--obstacle distance at each step. For each run, we compute the final goal distance, the minimum environment distance, and the environment-collision steps. We define the environment-collision step count as
\begin{equation}
C_{\mathrm{env}}=\sum_{t=1}^{T}\mathbb{I}\left[d_{\min}^{\mathrm{env}}(t)<0\right],
\label{eq:collision-steps}
\end{equation}
where \(T\) is the episode length and \(\mathbb{I}[\cdot]\) is an indicator function.

\subsection{Adversarial Stress-Test Design}  

We implemented perception-level attacks on top of the nominal SPARK pipeline. Instead of changing the MuJoCo environment, we injected attacks into the safety-filter perception path by monkey-patching compute\_pairwise\_info inside the safety-index module. This allows us to corrupt the robot--obstacle pairwise information immediately before each filter computes its collision-avoidance constraints.

We tested two perception-level attacks. For perception noise, we added Gaussian noise to the perceived pairwise distances, with standard deviations ranging from 0.02 to 0.10. For sensor latency, we buffered recent pairwise information and returned stale data with delays up to 10 simulation steps, cloning only the numerical arrays needed for memory-safe buffering. Each attack run produced a \texttt{.npz} file over a 2000-step episode. We measured degradation using safety failure, defined by Eq.~\ref{eq:collision-steps}, and task inefficiency, defined as the mean arm--goal distance:
\begin{equation}
G_{\mathrm{mean}}=\frac{1}{T}\sum_{t=1}^{T} d_{\mathrm{goal}}(t),
\label{eq:mean-goal-distance}
\end{equation}
where \(d_{\mathrm{goal}}(t)\) is the arm--goal distance from \texttt{dist\_goal\_arm}. We report both metrics across nominal, low, medium, and high attack intensities. For obstacle crowding, we changed the number of obstacles to 5, 15, and 30 while keeping the same benchmark case and parsing pipeline.

\section{Experimental Results}
\label{sec:experiments}

\subsection{Baseline Replication} 

\begin{figure*}[t]
    \centering
    \begin{minipage}[t]{0.32\textwidth}
        \centering
        \includegraphics[width=\linewidth]{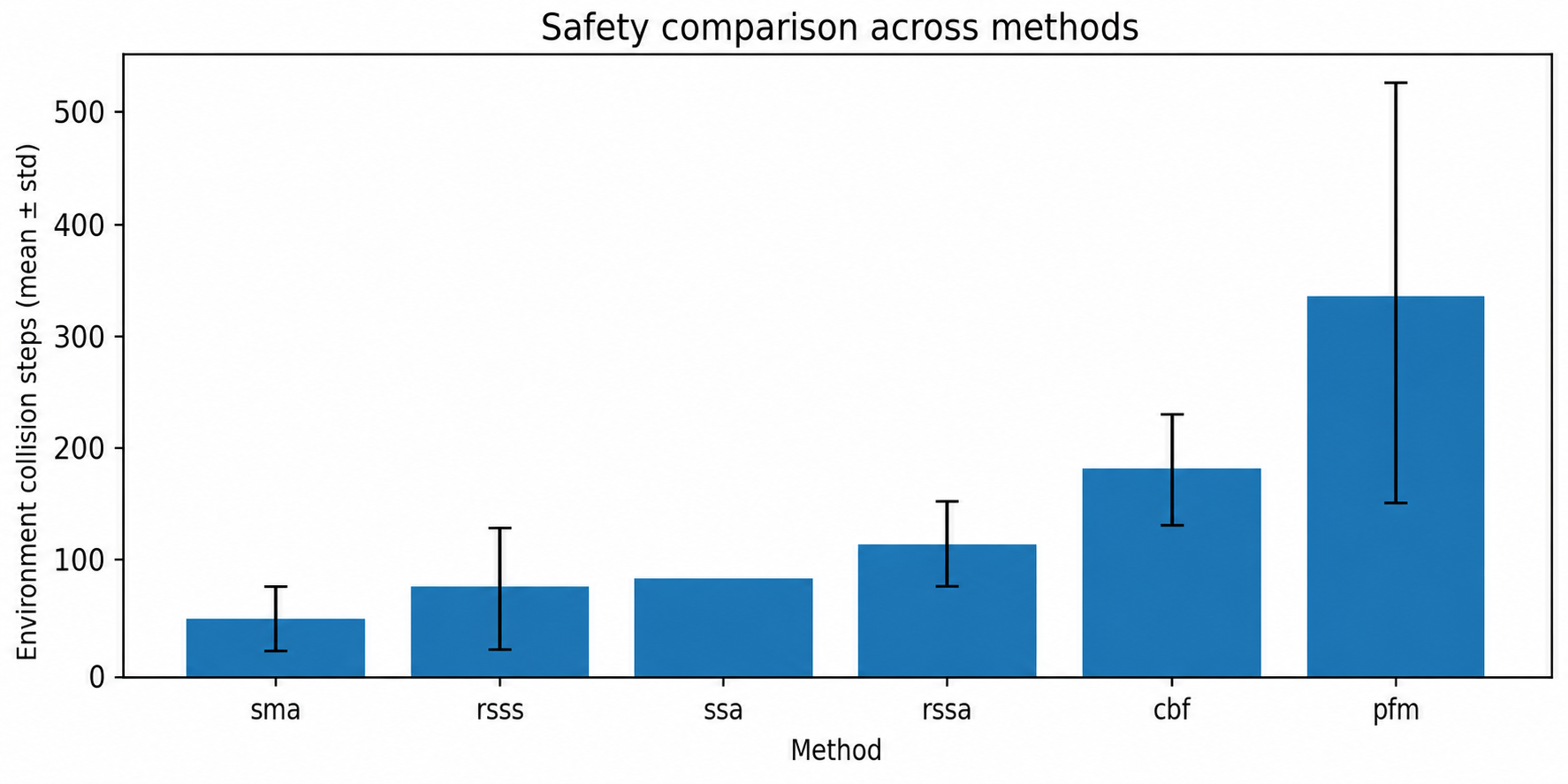}
        {\scriptsize (a) Environment collision steps}
    \end{minipage}
    \hfill
    \begin{minipage}[t]{0.32\textwidth}
        \centering
        \includegraphics[width=\linewidth]{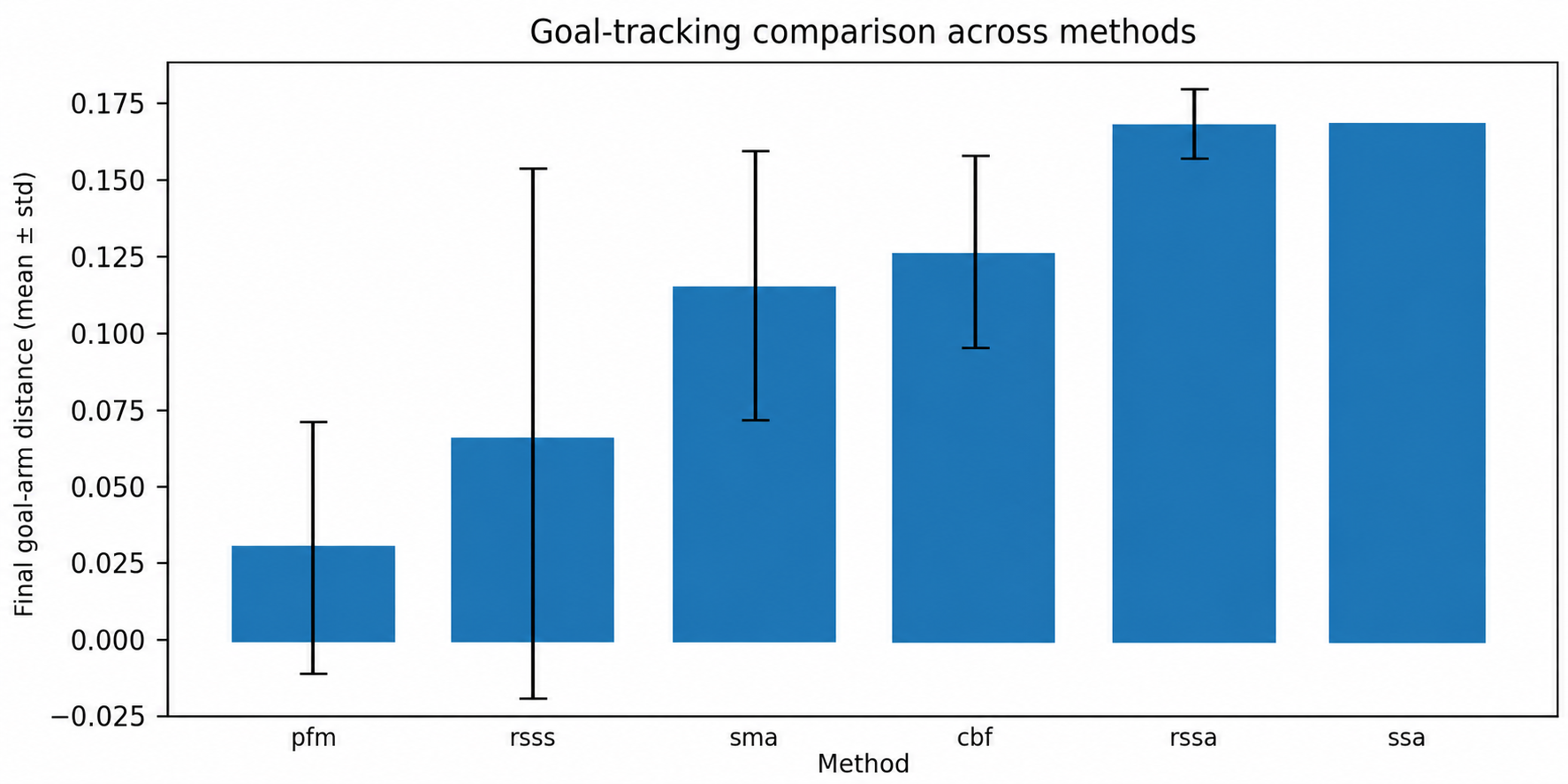}
        {\scriptsize (b) Final goal-arm distance}
    \end{minipage}
    \hfill
    \begin{minipage}[t]{0.32\textwidth}
        \centering
        \includegraphics[width=\linewidth]{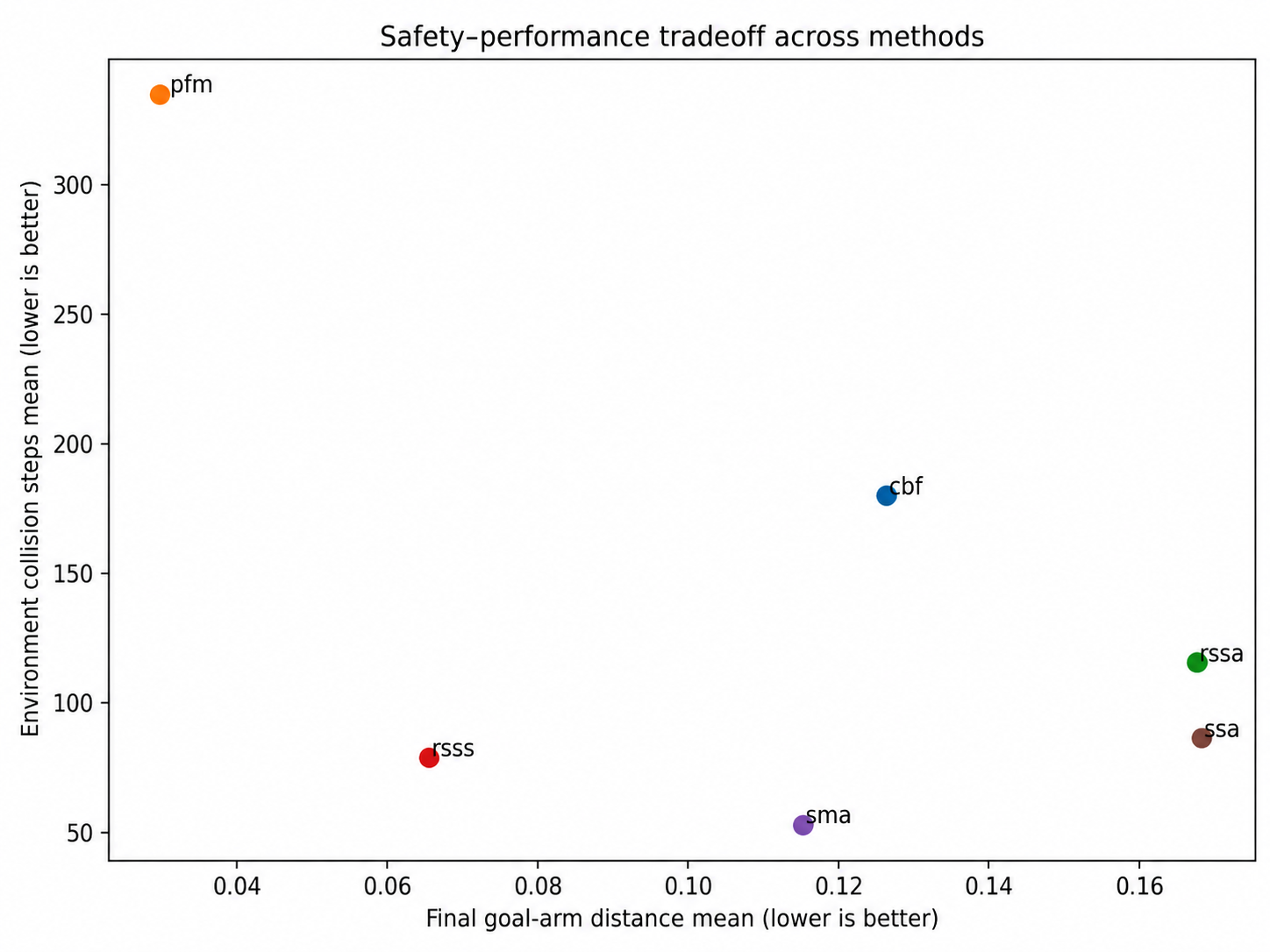}
        {\scriptsize (c) Safety--performance tradeoff}
    \end{minipage}

    \caption{Aggregate multi-seed comparison across six safety filters using seeds 20, 21, and 22 on \sparkcase{}. Lower values are better for both collision steps and final goal-arm distance.}
    \label{fig:aggregate_results}
\end{figure*}

We first generated diagnostic plots from the parsed SPARK metrics to verify that the pipeline correctly captured goal tracking, robot--environment distance, and self-distance. We then repeated the benchmark for seeds 20, 21, and 22 using the same case, \sparkcase{}, across RSSA, RSSS, SSA, CBF, PFM, and SMA. Figure~\ref{fig:aggregate_results} reports the mean and standard deviation across seeds. SMA has the fewest environment-collision steps, while PFM has the most. PFM most closely meets the arm goal, while SSA and RSSA remain farther from it. Overall, no method dominates both objectives: PFM favors goal tracking, SMA favors safety, and RSSS/ SSA show more balanced behavior.

\subsection{Obstacle-Crowding Stress Test}

We stress-test the nominal SPARK benchmark by increasing obstacle density while keeping the task and evaluation pipeline unchanged. Among the stress dimensions we consider, the completed experiments focus on obstacle crowding because it directly increases the number of robot--environment constraints.

\begin{figure}[t]
    \centering
    \includegraphics[width=\linewidth]{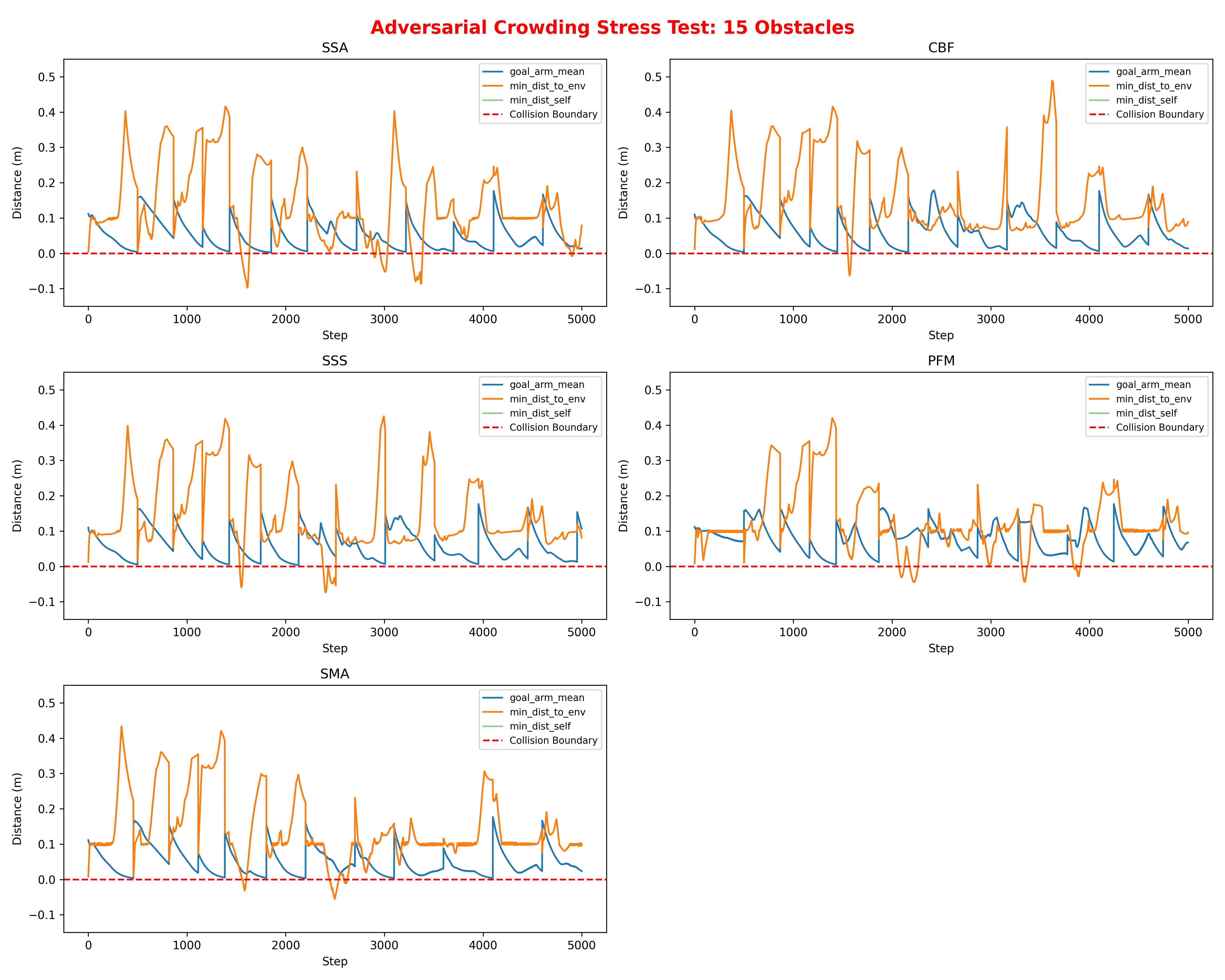}
    \caption{Time-series results for the 15-obstacle crowding stress test. The plot shows when each safety filter remains above the boundary and when it crosses into collision.}
    \label{fig:crowding-time-series-15obs}
\end{figure}

We run the same safety filters with 5, 15, and 30 obstacles. Figure~\ref{fig:crowding-time-series-15obs} shows the 15-obstacle case, while Table~\ref{tab:stress-test-results} summarizes all three obstacle-density settings. For each run, we track goal-arm distance, minimum robot--environment distance, minimum self-distance, and collision-boundary crossings. A negative environment distance indicates a collision.

We summarize each crowding level using two metrics: task inefficiency, measured by the mean final goal-arm distance, and safety failure, measured by the total number of environment collisions. Lower values are better for both.

\begin{table}[h]
\caption{Adversarial crowding results across obstacle-density settings. Lower values are better.}
\label{tab:stress-test-results}
\centering
\footnotesize
\setlength{\tabcolsep}{3.5pt}
\renewcommand{\arraystretch}{0.88}
\begin{tabular*}{\linewidth}{@{\extracolsep{\fill}}lcccccc@{}}
\toprule
& \multicolumn{2}{c}{5 obstacles} 
& \multicolumn{2}{c}{15 obstacles} 
& \multicolumn{2}{c}{30 obstacles} \\
\cmidrule(lr){2-3}\cmidrule(lr){4-5}\cmidrule(lr){6-7}
Method 
& Coll. $\downarrow$ & Goal $\downarrow$
& Coll. $\downarrow$ & Goal $\downarrow$
& Coll. $\downarrow$ & Goal $\downarrow$ \\
\midrule
CBF & \textbf{0} & \textbf{0.046} & \textbf{35} & 0.060 & 106 & 0.091 \\
PFM & \textbf{0} & 0.062 & 258 & 0.079 & 646 & 0.143 \\
SMA & \textbf{0} & 0.048 & 168 & \textbf{0.050} & \textbf{96} & \textbf{0.068} \\
SSS & \textbf{0} & 0.047 & 164 & 0.058 & 136 & 0.093 \\
SSA & 166 & 0.047 & 314 & \textbf{0.050} & 294 & 0.074 \\
\bottomrule
\end{tabular*}
\end{table}

Table~\ref{tab:stress-test-results} summarizes the 5-, 15-, and 30-obstacle crowding results. The safest method changes with obstacle density: most methods avoid collisions with 5 obstacles, CBF has the fewest collision steps with 15 obstacles, and SMA performs best with 30 obstacles. PFM degrades most strongly in the densest setting. Overall, the results show that better goal tracking does not always imply safer behavior under crowding. The 5- and 30-obstacle settings are summarized in the table to avoid duplicating similar time-series plots.

\begin{figure}[h]
    \centering
    \includegraphics[width=\linewidth]{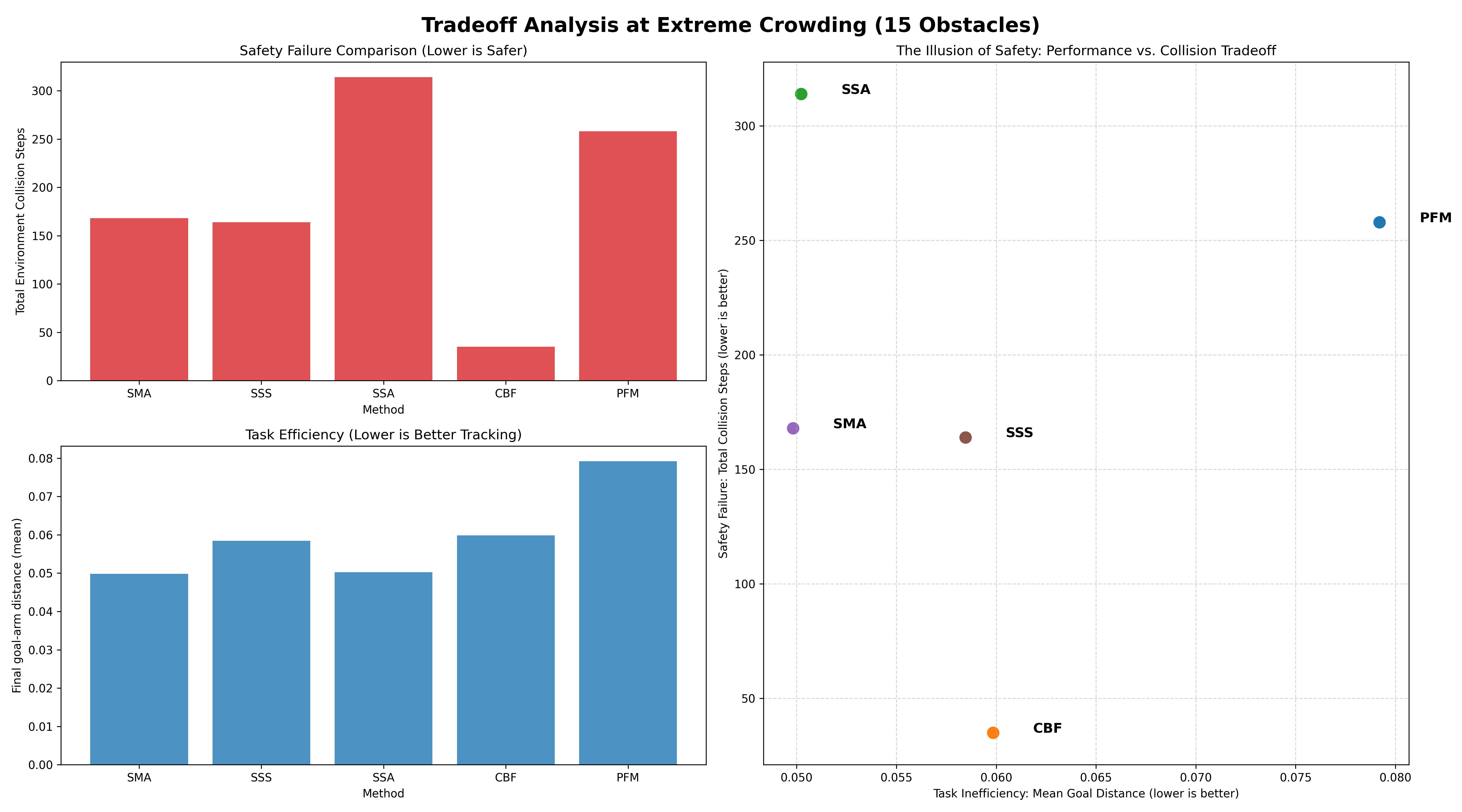}
    \caption{Trade-off summary for the 15-obstacle crowding stress test. Bars show collision steps and mean goal-arm distance; the scatter plot shows task inefficiency versus safety failure. Lower is better on all metrics.}
    \label{fig:crowding-tradeoff-15obs}
\end{figure}

\subsection{Perception Noise and Sensor Latency} 

We evaluate two perception-level attacks: Gaussian noise on perceived distances and latency in obstacle updates. These attacks keep the MuJoCo state unchanged, but corrupt or delay the information used by the safety filter.

\begin{figure}[t]
    \centering
    \includegraphics[width=\columnwidth]{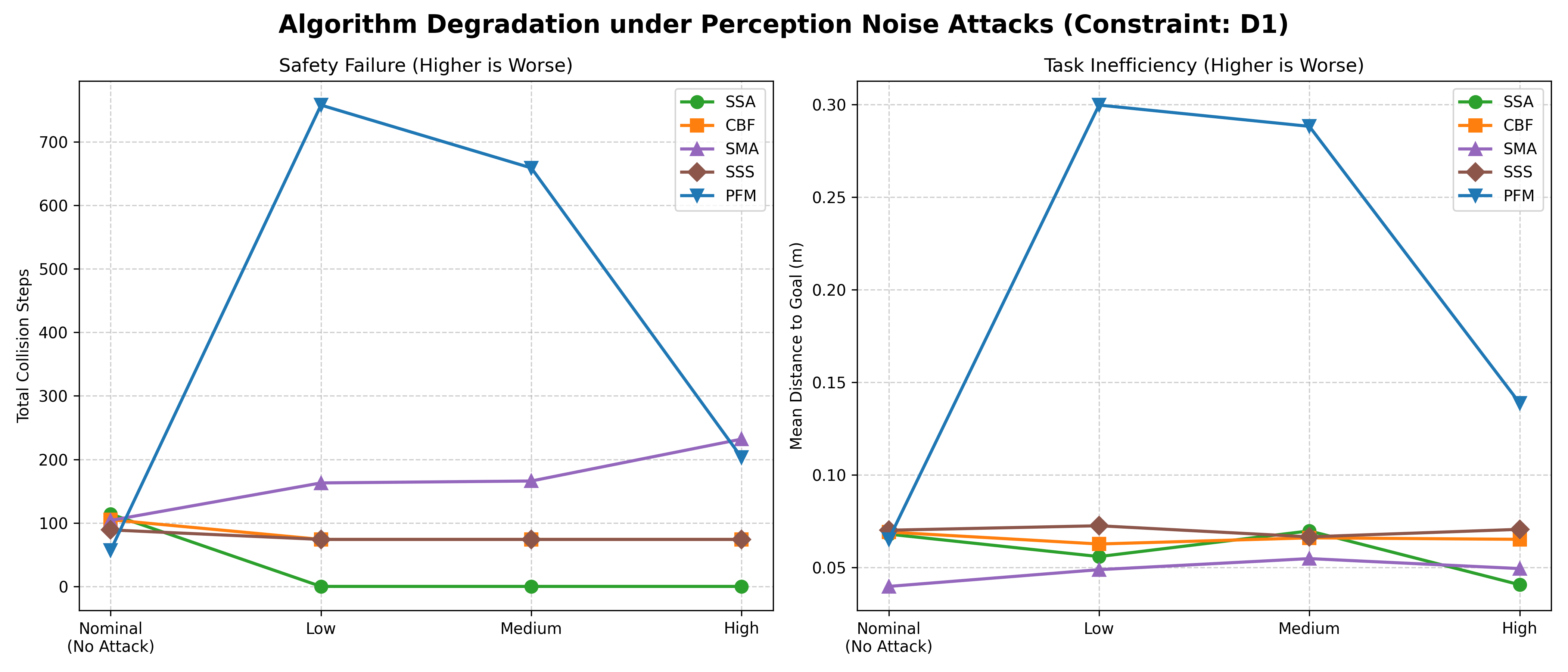}
    \caption{Algorithm degradation under perception-noise attacks. Safety failure is measured by total collision steps and task inefficiency is measured by mean arm-goal distance. The x-axis shows attack intensity from nominal to high; higher values indicate worse outcomes.}
    \label{fig:perception-noise-degradation}
\end{figure}

\begin{figure}[t]
    \centering
    \includegraphics[width=\columnwidth]{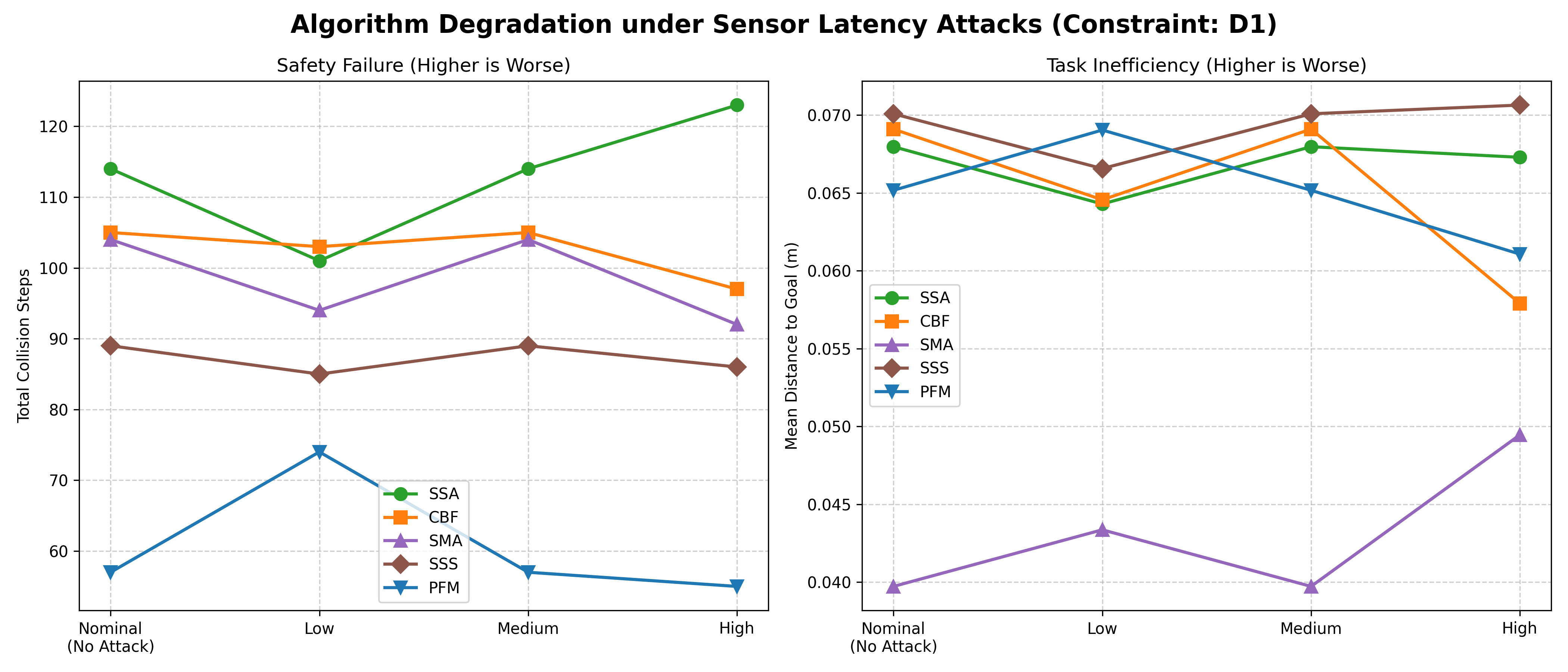}
    \caption{Algorithm degradation under sensor-latency attacks. Safety failure is measured by total collision steps and task inefficiency is measured by mean arm-goal distance. The x-axis shows attack intensity from nominal to high; higher values indicate worse outcomes.}
    \label{fig:latency-degradation}
\end{figure}

Figures~\ref{fig:perception-noise-degradation} and~\ref{fig:latency-degradation} show that the two attacks affect the filters differently. Under perception noise, PFM degrades sharply at low and medium intensity, with large increases in collision steps and goal error. SSA remains safer after the nominal setting, while CBF and SSS stay relatively stable; SMA shows moderate safety degradation as intensity increases. Under latency, collision counts are more similar across methods, but stale obstacle information still affects both safety and tracking behavior. PFM has the lowest collision count, whereas SSA has the highest at high latency. These results are preliminary, but they show that clean nominal performance does not fully characterize robustness under corrupted or delayed sensing.

\begin{figure}[b]
    \centering
    \includegraphics[width=\linewidth]{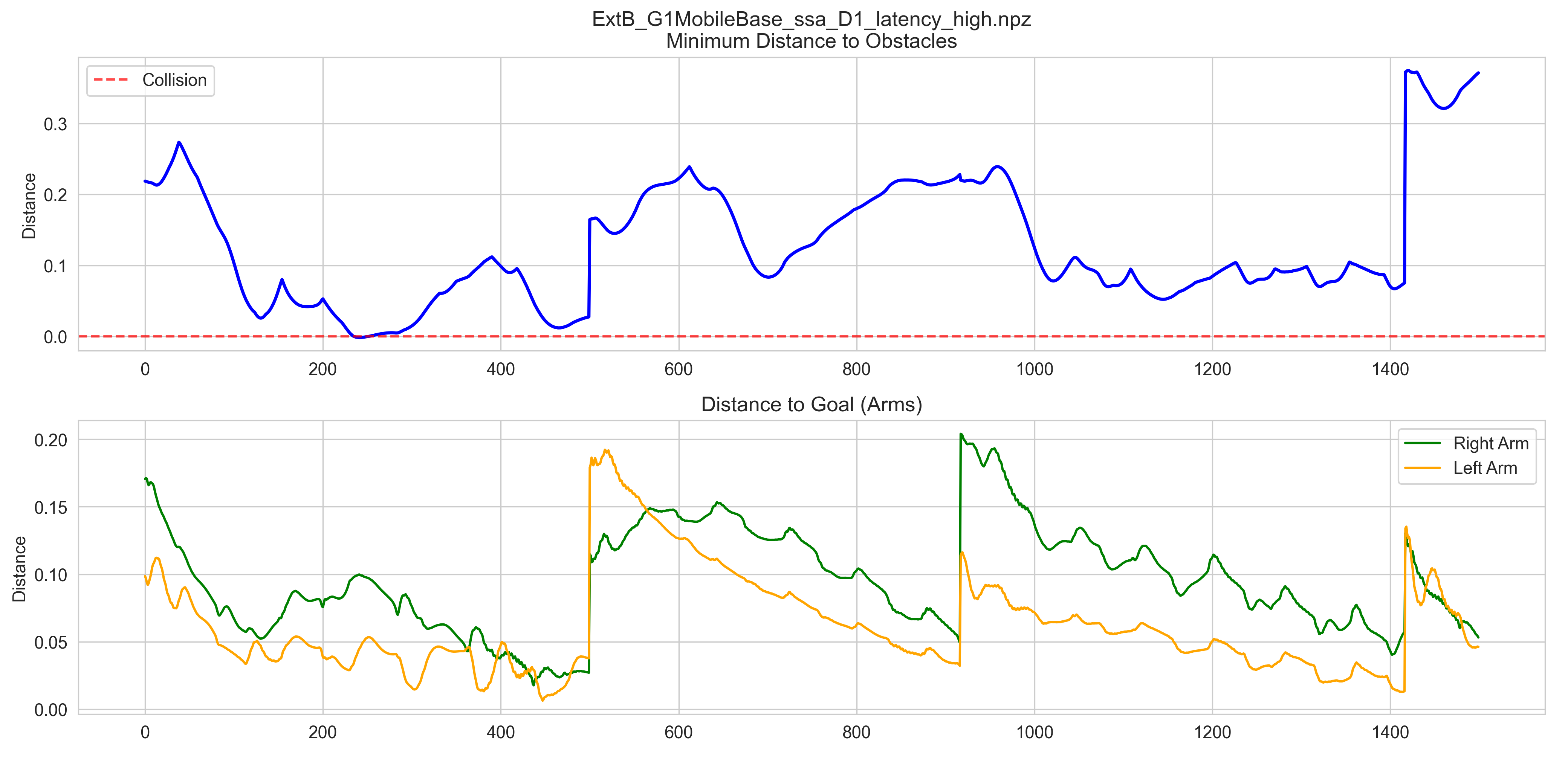}
    \caption{Diagnostic plot for SSA under high sensor latency. The top plot shows the minimum robot--environment distance with the red dashed collision boundary. The bottom plot shows right-arm and left-arm goal distances.}
    \label{fig:ssa-latency-high-diagnostic}
\end{figure}

Figure~\ref{fig:ssa-latency-high-diagnostic} shows a representative step-wise trace for SSA under high sensor latency. It illustrates how the minimum robot--environment distance and arm-goal distances evolve during one attack run.

\section{Discussion}
\label{sec:discussion}

One of the primary outcomes of this work is the development of an end-to-end SPARK replication and analysis pipeline. We ran the MuJoCo-based G1 benchmark locally, collected logs, and parsed high-dimensional \texttt{.npz} outputs, and converted them into interpretable safety metrics.

The main limitations are runtime cost and incomplete failure diagnostics. Full benchmark runs were slow, often taking several hours per seed, which limited the number of seeds and stress-test settings that could be run. We focused mainly on one SPARK case, \sparkcase{}, so the results should not be treated as broad claims across all SPARK tasks. Instead, the results should be read as evidence from a focused replication and robustness study. Some runs also produced repeated ``No Solution'' messages, suggesting feasibility issues, but we did not systematically measure the frequency of infeasibility.

\section{Conclusion and Future Work}
\label{sec:conclusion}

We replicated and analyzed the SPARK humanoid benchmark case \sparkcase{} using multiple safety filters and a common evaluation pipeline. We also built a post-processing workflow that converts raw \texttt{.npz} logs into final goal-arm distance, minimum robot--environment distance, and environment-collision steps. The results show a safety--performance trade-off in which some methods track the goal better, while others reduce collision steps more effectively. The stress tests further indicate that obstacle crowding, perception noise, and sensor latency can expose robustness issues that are not visible from nominal performance alone.

Future work should expand the study across more SPARK benchmark cases, more random seeds, and more systematic stress-test settings. A stronger evaluation should directly compare reproduced results against the original SPARK metrics, include constraint-conflict scenarios, and measure recovery time after unsafe events.

\section*{Author Contributions}

Saurav Ghosh led the overall study, including SPARK replication setup, experiment execution, data parsing, metric design, plotting, figure organization, and writing of the original manuscript draft. Abdou Sow contributed to the adversarial stress-test implementation and the perception-noise and sensor-latency analysis. Luke Zhang contributed to the diagnostic analysis for the SSA high-latency experiment.

\section*{Code Availability}

The code, configurations, parsing scripts, and plotting tools used in this study are available at:
\href{https://github.com/ghoshsaurav/spark-adversarial-safety}{https://github.com/ghoshsaurav/spark-adversarial-safety}.

%----------------------------------------------------------

\printbibliography

@misc {sun2025, title={SPARK: Safe Protective and Assistive Robot Kit}, url={https://arxiv.org/abs/2502.03132}, journal={arXiv.org}, author={Sun, Yifan and Chen, Rui and Yun, Kai S and Fang, Yikuan and Jung, Sebin and Li, Feihan and Li, Bowei and Zhao, Weiye and Liu, Changliu}, year={2025} }

@article{khatib1986real, title={Real-Time Obstacle Avoidance for Manipulators and Mobile Robots}, volume={5}, url={https://journals.sagepub.com/doi/abs/10.1177/027836498600500106}, DOI={10.1177/027836498600500106}, number={1}, journal={The International Journal of Robotics Research}, author={Khatib, Oussama}, year={1986}, month={Mar}, pages={90–98}}

@ARTICLE{gracia2013reactive,
  author={Gracia, Luis and Garelli, Fabricio and Sala, Antonio},
  journal={IEEE Transactions on Control Systems Technology}, 
  title={Reactive Sliding-Mode Algorithm for Collision Avoidance in Robotic Systems}, 
  year={2013},
  volume={21},
  number={6},
  pages={2391-2399},
  doi={10.1109/TCST.2012.2231866}}

@inbook{ames2019control, place={Piscataway, NJ}, title={Control Barrier Functions: Theory and Applications}, ISBN={978-3-907144-00-8}, DOI={10.23919/ECC.2019.8796030}, abstractNote={This paper provides an introduction and overview of recent work on control barrier functions and their use to verify and enforce safety properties in the context of (optimization based) safety-critical controllers. We survey the main technical results and discuss applications to several domains including robotic systems.}, booktitle={2019 18th European Control Conference (ECC)}, publisher={IEEE}, author={Ames, Aaron D. and Coogan, Samuel and Egerstedt, Magnus and Notomista, Gennaro and Sreenath, Koushil and Tabuada, Paulo}, year={2019}, month={Jun}, pages={3420–3431} }

@inproceedings{wei2019safe,
author = {Wei, Tianhao and Liu, Changliu},
title = {Safe Control Algorithms Using Energy Functions: A Uni ed Framework, Benchmark, and New Directions},
year = {2019},
publisher = {IEEE Press},
url = {https://doi.org/10.1109/CDC40024.2019.9029720},
doi = {10.1109/CDC40024.2019.9029720},
booktitle = {2019 IEEE 58th Conference on Decision and Control (CDC)},
pages = {238–243},
numpages = {6},
location = {Nice, France}
}

@article{nguyen2015safety,
title = {Safety-Critical Control for Dynamical Bipedal Walking with Precise Footstep Placement**This work is partially supported through funding from the Google Faculty Award and NSF Grant IIS-1464337.},
journal = {IFAC-PapersOnLine},
volume = {48},
number = {27},
pages = {147-154},
year = {2015},
note = {Analysis and Design of Hybrid Systems ADHS},
issn = {2405-8963},
doi = {https://doi.org/10.1016/j.ifacol.2015.11.167},
url = {https://www.sciencedirect.com/science/article/pii/S2405896315024258},
author = {Quan Nguyen and Koushil Sreenath}
}

@inproceedings{agrawal2017discrete,
author = {Agrawal, Ayush and Sreenath, Koushil},
year = {2017},
month = {07},
pages = {},
title = {Discrete Control Barrier Functions for Safety-Critical Control of Discrete Systems with Application to Bipedal Robot Navigation},
doi = {10.15607/RSS.2017.XIII.073}
}

@INPROCEEDINGS{khazoom2022humanoid,
  author={Khazoom, Charles and Gonzalez-Diaz, Daniel and Ding, Yanran and Kim, Sangbae},
  booktitle={2022 IEEE-RAS 21st International Conference on Humanoid Robots (Humanoids)}, 
  title={Humanoid Self-Collision Avoidance Using Whole-Body Control with Control Barrier Functions}, 
  year={2022},
  volume={},
  number={},
  pages={558-565},
  doi={10.1109/Humanoids53995.2022.10000235}
}

@article{paredes2024safe,
  title={Safe Whole-Body Task Space Control for Humanoid Robots},
  author={Victor Paredes and Ayonga Hereid},
  journal={2024 American Control Conference (ACC)},
  year={2023},
  pages={949-956},
  url={https://api.semanticscholar.org/CorpusID:265213414}
}

@article{brunke2022safe,
   author = "Brunke, Lukas and Greeff, Melissa and Hall, Adam W. and Yuan, Zhaocong and Zhou, Siqi and Panerati, Jacopo and Schoellig, Angela P.",
   title = "Safe Learning in Robotics: From Learning-Based Control to Safe Reinforcement Learning", 
   journal= "Annual Review of Control, Robotics, and Autonomous Systems",
   year = "2022",
   volume = "5",
   number = "Volume 5, 2022",
   pages = "411-444",
   doi = "https://doi.org/10.1146/annurev-control-042920-020211",
   url = "https://www.annualreviews.org/content/journals/10.1146/annurev-control-042920-020211",
   publisher = "Annual Reviews",
   issn = "2573-5144",
   type = "Journal Article",
}

@article{dawson2023safe,
author = {Dawson, Charles and Gao, Sicun and Fan, Chuchu},
title = {Safe Control With Learned Certificates: A Survey of Neural Lyapunov, Barrier, and Contraction Methods for Robotics and Control},
year = {2023},
issue_date = {June 2023},
publisher = {IEEE Press},
volume = {39},
number = {3},
issn = {1552-3098},
url = {https://doi.org/10.1109/TRO.2022.3232542},
doi = {10.1109/TRO.2022.3232542},
journal = {Trans. Rob.},
month = jun,
pages = {1749–1767},
numpages = {19}
}

@article{yuan2022safecontrolgym,
  author={Yuan, Zhaocong and Hall, Adam W. and Zhou, Siqi and Brunke, Lukas and Greeff, Melissa and Panerati, Jacopo and Schoellig, Angela P.},
  journal={IEEE Robotics and Automation Letters},
  title={Safe-Control-Gym: A Unified Benchmark Suite for Safe Learning-Based Control and Reinforcement Learning in Robotics},
  year={2022},
  volume={7},
  number={4},
  pages={11142-11149},
  doi={10.1109/LRA.2022.3196132}}

@inproceedings{ji2023safetygymnasium,
author = {Ji, Jiaming and Zhang, Borong and Zhou, Jiayi and Pan, Xuehai and Huang, Weidong and Sun, Ruiyang and Geng, Yiran and Zhong, Yifan and Dai, Juntao and Yang, Yaodong},
title = {Safety-Gymnasium: a unified safe reinforcemei learning benchmark},
year = {2023},
publisher = {Curran Associates Inc.},
address = {Red Hook, NY, USA},
booktitle = {Proceedings of the 37th International Conference on Neural Information Processing Systems},
articleno = {831},
numpages = {30},
location = {New Orleans, LA, USA},
series = {NIPS '23}
}

@InProceedings{donze2010breach,
author="Donz{\'e}, Alexandre",
editor="Touili, Tayssir
and Cook, Byron
and Jackson, Paul",
title="Breach, A Toolbox for Verification and Parameter Synthesis of Hybrid Systems",
booktitle="Computer Aided Verification",
year="2010",
publisher="Springer Berlin Heidelberg",
address="Berlin, Heidelberg",
pages="167--170",
isbn="978-3-642-14295-6"
}

@InProceedings{dreossi2019verifai,
author="Dreossi, Tommaso
and Fremont, Daniel J.
and Ghosh, Shromona
and Kim, Edward
and Ravanbakhsh, Hadi
and Vazquez-Chanlatte, Marcell
and Seshia, Sanjit A.",
editor="Dillig, Isil
and Tasiran, Serdar",
title="VerifAI: A Toolkit for the Formal Design and Analysis of Artificial Intelligence-Based Systems",
booktitle="Computer Aided Verification",
year="2019",
publisher="Springer International Publishing",
address="Cham",
pages="432--442",
isbn="978-3-030-25540-4"
}

@inproceedings{tuncali2018simulation,
  title = {Simulation-Based Adversarial Test Generation for Autonomous Vehicles with Machine Learning Components},
  author = {Tuncali, Cumhur Erkan and Fainekos, Georgios and Ito, Hisahiro and Kapinski, James},
  booktitle = {2018 IEEE Intelligent Vehicles Symposium (IV)},
  pages = {1555--1562},
  year = {2018},
  publisher = {IEEE},
  doi = {10.1109/IVS.2018.8500421}
}

@inproceedings{annpureddy2011s,
  title={S-taliro: A tool for temporal logic falsification for hybrid systems},
  author={Annpureddy, Yashwanth and Liu, Che and Fainekos, Georgios and Sankaranarayanan, Sriram},
  booktitle={International Conference on Tools and Algorithms for the Construction and Analysis of Systems},
  pages={254--257},
  year={2011},
  organization={Springer}
}

@misc{koren2018adaptive, title={Adaptive stress testing for autonomous vehicles}, url={https://arxiv.org/abs/1902.01909}, journal={arXiv.org}, author={Koren, Mark and Alsaif, Saud and Lee, Ritchie and Kochenderfer, Mykel J.}, year={2019}, month=feb }

@inproceedings{10.1145/3314221.3314633,
author = {Fremont, Daniel J. and Dreossi, Tommaso and Ghosh, Shromona and Yue, Xiangyu and Sangiovanni-Vincentelli, Alberto L. and Seshia, Sanjit A.},
title = {Scenic: a language for scenario specification and scene generation},
year = {2019},
isbn = {9781450367127},
publisher = {Association for Computing Machinery},
address = {New York, NY, USA},
url = {https://doi.org/10.1145/3314221.3314633},
doi = {10.1145/3314221.3314633},
booktitle = {Proceedings of the 40th ACM SIGPLAN Conference on Programming Language Design and Implementation},
pages = {63–78},
numpages = {16},
location = {Phoenix, AZ, USA},
series = {PLDI 2019}
}

@article{Chen2018HamiltonJacobiRS,
  title={Hamilton-Jacobi Reachability: Some Recent Theoretical Advances and Applications in Unmanned Airspace Management},
  author={Mo Chen and Claire J. Tomlin},
  journal={Annu. Rev. Control. Robotics Auton. Syst.},
  year={2018},
  volume={1},
  pages={333-358},
  url={https://api.semanticscholar.org/CorpusID:262693302}
}

@proceedings{liu2014safe,
    author = {Liu, Changliu and Tomizuka, Masayoshi},
    title = {Control in a Safe Set: Addressing Safety in Human-Robot Interactions},
    volume = {Volume 3},
    series = {Dynamic Systems and Control Conference},
    pages = {V003T42A003},
    year = {2014},
    month = {10},
    doi = {10.1115/DSCC2014-6048},
    url = {https://doi.org/10.1115/DSCC2014-6048},
    eprint = {https://asmedigitalcollection.asme.org/DSCC/proceedings-pdf/DSCC2014/46209/V003T42A003/4446881/v003t42a003-dscc2014-6048.pdf},
}

%----------------------------------------------------------

\end{document}